\documentclass[10pt,twocolumn,letterpaper]{article}
\usepackage{cvpr}
\usepackage{times}
\usepackage{epsfig}
\usepackage{graphicx}
\usepackage{amsmath}
\usepackage{amssymb}
\usepackage[retainorgcmds]{IEEEtrantools}
\usepackage{algorithmic}
\usepackage{algorithm}
\usepackage{subfigure}

\usepackage{hyperref}
\usepackage[applemac]{inputenc}
\usepackage{pst-sigsys}
\usepackage{epstopdf}
\cvprfinalcopy % *** Uncomment this line for the final submission

\def\cvprPaperID{1456} % *** Enter the CVPR Paper ID here

\providecommand{\argmin}{\mathop{\textup{argmin}}}

\newcommand{\R}{\mathbb{R}}

\newcommand{\om}{\omega}
\newcommand{\ga}{\gamma}

\title{Classification with Scattering Operators}
\author{
Joan Bruna and St\'ephane Mallat\\
CMAP, Ecole Polytechnique, 91128 Palaiseau \\
{\it Proceedings of the IEEE CVPR 2011 conference}}

\date{March 2011}

\begin{document}
\maketitle
%\section{}
%\subsection{}

\begin{abstract}
A scattering vector is a local 
descriptor including 
multiscale and multi-direction co-occurrence
information. It is computed with 
a cascade of wavelet decompositions
and complex modulus. This scattering representation
is locally translation invariant
and linearizes deformations.
A supervised classification algorithm
is computed 
with a PCA model selection on scattering vectors. 
State of the art results
are obtained for handwritten digit recognition and
texture classification.\footnote{This work is funded by the ANR grant 0126 01.}
\end{abstract}

%\begin{keywords}
%Image classification, Invariant representations, local image descriptors, pattern recognition,
%texture classification.
%\end{keywords}

\section{Introduction}

%sift and similar descriptors can be cast as first order wavelet modulus descriptors
Locally invariant image descriptors such as
SIFT \cite{SIFT} provide efficient image representations for 
image classification and registration \cite{SIFT}.
These feature vectors as well as multiscale texture descriptors
can be computed with a spatial 
averaging of wavelet coefficient amplitudes.
The averaging reduces the feature variability and provides 
local translation invariance, but it also reduces information.

Scattering operators recover the lost high frequencies
and retransform them
into co-occurrence coefficients at multiple scales and orientations.
They provide much richer descriptors
of complex structures such as corners, junctions and multiscale texture variations.
These coefficients are locally translation invariant and they linearize
small deformations.
They are computed with a convolution network \cite{LeCun} which cascades
contractive wavelet transforms and modulus operators \cite{stephane1}.
Scattering operators provide new representations of stationary image
textures, which can discriminate texture having the same power spectrum.

The scattering transform of a class of signals is approximated
by an affine space computed with a PCA. Images are classified by selecting
a best approximation space model for their scattering transform.
State of the art results are obtained for hand-written digit recognition
and for texture discrimination, with important
rotation and illumination variability, and small training sets.

Section \ref{First} reviews the relations between wavelet transforms
and computer vision descriptors.
Section \ref{SOS} introduces scattering image representations.
Classification by scattering model selection
is introduced in Section \ref{numeric}, with numerical results.
Softwares are available at \texttt{ www.cmap.polytechnique.fr/scattering}.

\section{Scattering}

A scattering transform computes local image descriptors with
a cascade of wavelet decompositions, complex modulus
and a local averaging. The resulting scattering representation is
locally invariant to translations. 
It includes coefficients which are similar
to SIFT descriptors, together with 
co-occurrences coefficients 
at multiple scales and orientations.

\subsection{From Wavelets to SIFT and Textons}
\label{First}

Image feature vectors such as SIFT and multiscale Gabor
textons are obtained by averaging
the amplitude of wavelet coefficients, calculated with directional wavelets.
Writing these feature vectors as wavelet coefficients
helps to understand and to improve their properties.

Let $R_\ga x$ be the rotation of $x \in \R^2$ by an angle $\ga$.
Directional wavelets are obtained by rotating a single 
$\psi$, along $K$ angles $\gamma \in \Gamma$. Scaling them 
by $2^j$ yields
\[
\psi_{j,\gamma} (x) =  2^{-2j} \psi (2^{-j} R_{\gamma} x)~.
\]
The directional wavelet transform of $f$ at a position $x$ for
scales $2^j < 2^J$ is a vector of coefficients
\begin{equation}
\label{wavedfn}
W_J f(x) = \left(
\begin{array}{c}
f \star \psi_{j,\gamma} (x)\\
f \star \phi_J (x)
\end{array}
\right)_{j < J , \gamma \in \Gamma}
\end{equation}
where $\phi_J (x) = 2^{-2J} \phi(2^{-J} x)$ is a low-pass filter which
carries the low frequencies of $f$ above the scale $2^J$:
$\int \phi(x) dx = 1$.
Let $|W_J f(x)|^2$ be the Euclidean norm of this vector which sums the
square of its coordinates. Let $\hat f(\om)$ be the Fourier transform
of $f$. If wavelets satisfy 
\begin{equation}
\label{pars}
\sum_{j=-\infty}^{-1} \sum_{\gamma \in \Gamma} |\hat \psi_\ga (2^j \omega)|^2 +
|\hat \phi(\om)|^2 \leq 1
\end{equation}
then one can verify \cite{stephane1} that
\[
\|W_J f \|^2 = \int |W_J f(x)|^2 dx \leq \|f\|^2 = \int |f(x)|^2 \,dx~
\]
and this inequality is an equality if (\ref{pars}) is an equality.
The wavelet transform is then contractive and potentially unitary.

Many standard image feature vectors are obtained by averaging wavelet
coefficient amplitudes.
SIFT coefficients are obtained from histograms of image gradients
calculated at a fine scale $2^j$.  
A histogram bin indexed by $\gamma \in \Gamma$ stores the
local sum of the amplitudes of all gradient 
vectors whose orientations are 
close to $\gamma$. Several authors \cite{DAISY} 
observed that approximate SIFT feature vectors 
are computed more efficiently by averaging 
directly the partial derivative amplitudes of 
$f$ along the $K$ directions $\gamma \in \Gamma$,
with a low-pass filter $\phi_J$.
These averaged partial derivative amplitudes can be written
as averaged wavelet coefficients 
\[
|f \star \psi_{j,\gamma}| \star \phi_J (x)~,
\] 
with a partial derivative wavelet 
$\psi (x) = {\partial g(x)}/ {\partial x_1}$,
with $g(x) = e^{-|x|^2/2}$ and $x = (x_1,x_2)$. 
These averaged wavelet
coefficients are nearly invariant to translations or deformations
which are small relatively to $2^J$.

Partial derivative wavelets are well adapted to detect edge type elements,
but these wavelets
do not have enough frequency and directional resolution 
to discriminate more complex structures appearing in textures.
For texture analysis, wavelets
with a better frequency localization are often used \cite{Malik}.
Complex Gabor functions are examples of such directional
wavelets obtained by modulating
a Gaussian window at a frequency $\xi$:
\begin{equation}
\label{Gabor}
\psi (x) = e^{i \xi x_1} \,e^{-|x|^2 / 2}~.
\end{equation}
For stationary textures, 
$|f \star \psi_{j,\gamma}| \star \phi_J (x)$ 
has a reduced stochastic variability
because of the averaging kernel $\phi_J$.

\subsection{Scattering Coefficients}
\label{SOS}

The local translation invariance and variability reduction 
of SIFT descriptors and multiscale textons is
obtained by averaging.
Scattering operators restore part
of the information lost by this averaging
with co-occurrence
coefficients having similar invariance properties.

The wavelet transform (\ref{wavedfn}) shows that high frequencies eliminated
in $|f \star \psi_{j_1,\ga_1}| \star \phi_J$ 
by the convolution with $\phi_J$ are recovered by convolutions with
wavelets $|f \star \psi_{j_1,\ga_1}| 
\star \psi_{j_2,\ga_2}$ at scales $2^{j_2} < 2^{J}$.
To become insensitive to local translation and reduce the variability
of these coefficients, their complex phase is removed by a modulus, and it is
averaged by $\phi_J$:
\[
||f \star \psi_{j_1,\ga_1}| \star \psi_{j_2,\ga_2}| \star \phi_J~.
\]
These are called scattering coefficients because they result from
all interferences of $f$ with two wavelets \cite{stephane0}.
They give co-occurrence information in $f$ for any pair of 
scales $2^{j_1}$, $2^{j_2}$ and any two directions ${\ga_1}$ and
$\ga_2$. This can distinguish corners and junctions from edges and 
it characterizes texture structures.
Coefficients are only calculated for $2^{j_2} < 2^{j_1}$ because one can show
\cite{stephane1} that 
$|f \star \psi_{j_1,\ga_1}|\star \psi_{j_2,\ga_2}$
is negligible at scales $2^{j_2} \geq 2^{j_1}$. 

The convolution with $\phi_J$ removes high frequencies 
and thus yields
second order coefficients that are  locally translation invariant.
High frequencies can again be restored by
finer scale wavelet coefficients, which 
are regularized by averaging their amplitude 
with $\phi_J$. Applying iteratively this procedure
$q$ times yields a vector of coefficients at each $x$:
\[
S_{q,J} f(x)=
\Bigl(
|||f \star \psi_{j_1,\ga_1}| \star ... \star| \psi_{j_q,\ga_q}|\star \phi_J(x) \Bigr)_{j_1 < ...<j_q < J \atop (\ga_1,...,\ga_q) \in \Gamma^q}
\]
This vector has $K^q\binom{J}{q}$ scattering coefficients,
computing interactions 
between  $f$ and the successive wavelets $\psi_{j_1,\ga_1}\,$...$\,\psi_{j_q,\ga_q}$.
A scattering vector aggregates all these coefficients
up to a maximum order $q \leq m$:
\[
S_J f (x) = 
\Bigl( S_{q,J} f(x) \Bigr)_{0 \leq q \leq m}~,
\]
and the first coefficient is the signal average
$S_{0,J} f (x) =f \star \phi_J (x)$. 
The scattering vector size is $\sum_{q=0}^m K^q\binom{J}{q}$.
After convolution with $\phi_J$ the output is subsampled at intervals
$2^J$. If $f(n)$
is an image of $N$ pixels, this uniform sampling yields a scattering 
representation $S_J f (2^J n)$ including a total of 
$N_J = 2^{-2J} N \sum_{q=0}^m K^q\binom{J}{q}$ coefficients.

A scattering vector is computed with a cascade of convolutions
and modulus operators over $m+1$ layers,
like in convolution network architectures \cite{LeCun,Poggio}: 
\begin{equation*}
\begin{matrix}
f(n) & \rightarrow & 
%\boxed
{f \star \phi_J(2^J n)} \\
\downarrow & & \\
|f \star \psi_{j_1,\ga_1} | & \rightarrow & 
%\boxed
{|f \star \psi_{j_1,\ga_1}| \star \phi_J (2^J n)} \\
\downarrow & & \\
||f \star \psi_{j_1,\ga_1}| \star \psi_{j_2,\ga_2} | & \rightarrow & 
%\boxed
{ ||f \star \psi_{j_1,\ga_1}| \star \psi_{j_2,\ga_2}| \star \phi_J (2^J n)} \\
\downarrow & & \\
... & & 
\end{matrix}
\end{equation*}
To reduce computations, wavelet convolutions are subsampled
at intervals proportional to the last scale $2^{j_q}$, with an oversampling
factor of $2$:
\[
|||f \star \psi_{j_1,\ga_1}| \star ... \star| \psi_{j_q,\ga_q}(2^{j_q-1} n)|~.
\]
A final low-pass filtering and subsampling yields
\[
|||f \star \psi_{j_1,\ga_1}| \star ... \star| \psi_{j_q,\ga_q}|\star \phi_J(2^J n)
\]
With an FFT, 
the overall computational complexity is then $O(N \log N)$.

\subsection{Scattering Distance and Deformation Stability}

The scattering transform defines a distance between two images $f$ and $g$.
This distance has important invariance and stability properties 
that are briefly reviewed.
Let $|S_J f(x)|^2$ be the squared Euclidean norm of
the vector $S_J f(x)$. The scattering distance of $f$ and $g$ is
\begin{equation}\label{scatdistance}
\|S_J f - S_J g \|^2 = \int |S_J f(x) - S_J g(x)|^2\,dx .
\end{equation}
For discrete images,
the integral is replaced by a discrete sum. The scattering operator
$S_J$ is contractive because it is 
a cascade of wavelet transforms $W_J$ 
and modulus operators, which are both contractive \cite{slotine}:
\[
\|S_J f - S_J g \|^2 \leq \|f-g\|^2 = \int |f(x) - g(x)|^2\,dx~.
\]
In particular $\|S_J f \|^2 \leq \|f\|^2$. 
If the maximum order is $m = \infty$ then 
one can prove \cite{stephane1} that if the wavelet transform is unitary
then for appropriate
complex wavelets $\|S_J f \|^2 = \|f\|^2$.
The energy of $f$ is thus spread across scattering coefficients of
multiple orders, but this energy has a fast decay
as the co-occurrence order $q$ increases. In the Caltech101 image database,
98\% of the energy $\|S_J f\|^2$ is carried by scattering coefficients of order
$0$, $1$ and $2$. In applications, we shall thus limit the scattering order
to $m=2$. The energy of all scattering coefficients of order $2$, 
$||f \star \psi_{j_1,\ga_1}| \star \psi_{j_2,\ga_2}| \star \phi_J$,
is about 20\% of the energy of all order $1$ coefficients
$|f \star \psi_{j_1,\ga_1}| \star \phi_J$, which is not negligible.
We shall see that order 2 coefficients have indeed 
an important impact on classification results.

The efficiency of a scattering representation comes from its
invariance to local translations due to convolutions with $\phi_J$,
and from its ability to linearize deformations.
Let $D_\tau f(x) = f(x-\tau(x))$ be a deformation
of $f$ with a regular displacement field  $\tau(x)$.
It is a pure translation only if $\nabla \tau =0$.
We write $|\tau|_\infty = \sup_x |\tau (x)|$ the maximum translation amplitude,
and $|\nabla \tau|_\infty = \sup_x |\nabla \tau (x)|$ the maximum deformation
amplitude,
where $|\nabla \tau (x)|$ is the matrix sup norm of $\nabla \tau (x)$.
The sup-norm of the Hessian of $\tau$ is also written $|H\tau|_\infty$.
It is shown in \cite{stephane1} that the scattering metric
satisfies
\begin{equation}\label{lipschitz}
\|S_{J} (D_\tau f) - S_{J}f\| \leq C m \|f\| 
\Big(2^{-J} |\tau|_\infty + J (|\nabla \tau|_\infty + |H\tau|_\infty )\Big).
\end{equation}
The first term $2^{-J} |\tau|_\infty$ is the translation error which
is small if $2^J \gg |\tau|_\infty$. The other terms are dominated by the
deformation amplitude $|\nabla \tau|_\infty$.
If $2^J \geq |\tau|_\infty / |\nabla \tau|_\infty$ then
two deformed signals have a scattering distance essentially
proportional to the deformation amplitude $|\nabla \tau|_\infty$.

\section{Classification by Affine Model Selection}
\label{numeric}

A scattering representation $S_J f$ is
invariant to small translations
relatively to $2^J$. It linearizes deformations and provides co-occurence
descriptors. A classifier
is obtained by selecting an affine space model 
which best approximates $S_J f$.

Each signal class is represented
by a random vector $F_i$ whose realizations are images of $N$ pixels
in the class. Scattering vectors $S_J F_i (2^J n)$ define an image
representation with a total of
$N_J = 2^{-2J} N \sum_{q=0}^m K^q\binom{J}{q}$ coefficients.
Let $E\{S_J F_i (2^J n)\}$ be their expected values.
Deformations of $F_i$ are mostly linearized by $S_J$ and thus produce a
variability $S_J F_i - E\{S_J F_i\}$ 
which is well approximated in a linear space of low dimension $d$.
This linear space is computed with a PCA by diagonalizing
the covariance  of $S_J F_i$. We denote by
${\mathbf{V}_{d,i}}$ the space generated by the $d$ 
covariance eigenvectors of largest variance. The dimension $d$ is
adjusted so that $S_J F_i$ is closely 
approximated by its projection in the affine space
\[
{\bf A}_{d,i} = E \{S_J F_i\} + {\bf V}_{d,i} ~.
\]
in comparison with the error produced by the affine spaces
${\bf A}_{d,i'}$, $i' \neq i$, corresponding to the other classes.

A signal $f$ will be 
associated to the class $\hat \i$ which yields the best affine space
approximation:
\begin{equation}
\label{clasdifnsdf}
\hat{\i}(f) = \argmin_{i \leq I}
\|S_J f - P_{\mathbf{A}_{d,i}}(S_J f) \|~.
\end{equation}
Observe that
\[
\|S_J f - P_{\mathbf{A}_{d,i}}(S_J f) \| = 
\|P_{\mathbf{V}_{d,i}^\perp} (S_J f - E \{S_{J} F_i\}) \|
\]
where $\mathbf{V}_{d,i}^\perp$ is the orthogonal complement of 
$\mathbf{V}_{d,i}$. Minimizing the affine space approximation error is thus
equivalent to minimize the distance between $S_J f$ and the
class centroid $E\{S_J F_i\}$, without taking into account the first $d$ 
principal variability directions. 
A cross-validation procedure 
finds the dimension $d$ and the scale $2^J$ 
which yields the smallest classification
error. This error is computed on a subset of the training images 
that is not used for the PCA calculations.

Affine space scattering models can be interpreted
as generative models computed independently for each class. 
As opposed to
discriminative classifiers such as an SVM, no 
interaction between classes is taken into account, besides the
choice of the model dimensionality $d$.

Classification results are given for hand-written digits and textures
that are deformed, rotated, scaled and have illumination variations.
Scattering descriptors are computed 
with the complex Gabor wavelet (\ref{Gabor}) 
for $\xi = 3\pi / 4$, rotated along angles $k \pi/K$ with
$0 \leq k < K = 6$.
The lowpass filter is the 
Gaussian $\phi_J(x) = \lambda_J\exp(-(3 x/2^{J+1})^2/2)$
with $\int \phi_J (x) dx = 1$.

\subsection{Handwritten digit recognition}

%DONNER LES TABLEAUX AVEC DIMENSIONS D'ESPACE ET LA COMPARAISON AVEC
%UN SVM. DONNER UNE INDICATION DE L'ORDRE DE GRANDEUR DES ERREURS 
%D'APPROXIMATION PAR PROJECTION DANS LES ESPACES AFFINE.

The MNIST database of hand-written digits is an example 
of structured pattern classification, where 
most of  the intra-class variability is due to
local translations and deformations. It comprises 
at most 60000 training samples and 10000 test samples. 
The state of the art 
is achieved with deep-learning convolutional networks
\cite{ranzato_cvpr} and dictionary learning \cite{mairal}.

Table \ref{MNIST} compares the scattering PCA classifier at maximum orders
$m=1$, $m=2$ and $m=3$. Cross validation finds an optimal scattering 
scale $2^J = 2^3$. This value is
compatible with observed deformations of digits whose amplitude
is typically at most $8$ pixels. For $J = 3$, there are
$N/64$ second order scattering vectors $S_J f$ of dimension $127$ each.

Below $5\, 10^3$ training samples, the scattering PCA classifier 
improves results of deep-learning convolutional networks.
For $m=2$, second order scattering coefficients
improve classification results obtained with $m=1$, but
a third order $m=3$ scattering yields marginal improvements. 
An SVM classifier is also 
applied on scattering vectors for $m=2$, with
a polynomial kernel whose degree was optimized. Minimum errors
are obtained with a degree $4$. The SVM error is well above
the PCA model selection error up to 60000 samples.
For small training sets,
it was indeed shown \cite{generative_vs_discriminative} 
that generative models, which do not estimate cross terms between classes,
can outperform discriminative classifiers such as SVM. 
 
\begin{table}[t]
\caption{Percentage of error as a function of the training size for MNIST,
for a Convolution Network \cite{ranzato_cvpr}, an SVM over scattering
coefficient for $m=2$, a PCA for $m=1,2,3$.
Minimum errors are in bold.} 
\label{MNIST}
\begin{center}
\begin{tabular}{|c | c c c c c|}
\hline
Training & Conv. & SVM & PCA & PCA & PCA \\ 
size  & Net. & $m=2$ &  $m=1$ & $m=2$ & $m=3$ \\ 
\hline
300 & $7.18$ & $21.5$ & $7.03$ & $6.05$ & $\bf{5.97}$ \\ %& 24\\
1000 & $3.21$ & $3.06$ & $2.99$ & $2.39$  & $\bf{2.37}$ \\%& 34\\
2000 & $2.53$ & $1.87$ & $2.11$ & $1.71$ & $\bf{1.71}$ \\%& 50\\
5000 &  $1.52$ & $1.54$ & $1.85$ & $1.57$ & $\bf{1.22}$ \\%& 40 \\
10000 & $\bf{0.85}$ & $1.15$ & $1.61$ & $1.17$ & $0.99$ \\%& 130 \\
20000 & $\bf{0.76}$ &  $0.92$ & $1.4$ & $0.96$ & $0.82$ \\%& 180 \\
40000 & $\bf{0.65}$ & $0.85$ & $1.32$ & $0.78$ & $0.79$ \\%& 180 \\
60000 & $\bf{0.53}$ & $0.7$ & $1.4$ & $0.77$ & $0.72$ \\%& 210\\
\hline
\end{tabular}
\end{center}
\end{table}

\begin{table}[t]
\caption{Values of the dimension $d$ of affine approximation models, of the 
intra class normalized approximation error $\sigma^2_d$,
and of the ratio $\lambda_d$ between inter class and intra class 
approximation errors, as a function of the training size.}
\label{MNIST_dimension}
\begin{center}
\begin{tabular}{|c | c c c |}
\hline
Training & $d$ & $\sigma^2_d$ & $\lambda_d$ \\ 
\hline
300 & $24$ & $2 \cdot 10^{-2}$ & $2.4$\\
5000 &  $40$ & $5 \cdot10^{-3}$ & $3.6$ \\
40000 & $180$ & $6\cdot 10^{-4}$ & $4.3 $ \\
\hline
\end{tabular}
\end{center}
\end{table}

Table \ref{MNIST_dimension} gives the dimension $d$ of affine
approximation spaces calculated by cross validation, for $m=2$.
The normalized approximation error $\sigma_d^2$ is
the expected approximation
error $E \{ \|S_J F_i - P_{\bf A_{i,d}} (S_J F_i) \|^2 \}$ in a class $i$
divided by the squared norm of $S_J F_i$, averaged over all $i$ 
and all $F_i$ in the test set.
Table \ref{MNIST_dimension} shows that the cross-validation
calculation of $d$ yields small approximation errors.
Table \ref{MNIST_dimension} also gives
the relative approximation error 
$$\lambda_d = \frac{E \{\min_{i'\neq i}  \|S_J F_i - P_{\bf A_{i',d}} (S_J F_i) \|^2\}}{E \{\|S_J F_i - P_{\bf A_{i,d}} (S_J F_i) \|^2\}}$$
produced by the closest affine model of a different class than that of $F_i$, averaged over all classes.
%$\lambda_d$ 
%$\|S_J f - P_{\bf A_{i,d}} (S_J f) \|^2$ when $f$ is not in the class $i$
%divided by the error when $f$ is in the class $i$, averaged over all classes.
As expected, when the training set increases, the dimension
$d$ increases so $\sigma_d^2$ decreases and the relative approximation error
$\lambda_d$ increases, which reduces the error rate.

Rotation invariance in the MNIST database is studied in the same setting
as in \cite{mnist_rotated}. The authors have constructed a 
transformed database with 12000 training samples and 50000 test images,
where samples are rotated versions of the digits using a uniform distribution
in $[0,2\pi]$. The PCA incorporates rotation invariance by increasing the 
dimension $d$ of the affine space ${\bf A}_{i,d}$. It removes
the main variability directions of $S_J f$ due to rotations. Error rates in
Table \ref{MNIST_rotated} are smaller with a scattering PCA than with
a convolution network \cite{mnist_rotated}. 
Better results are obtained with $m=2$ than with $m=1$ because
second order coefficients
maintain enough discriminability despite the removal of a larger
number $d$ of principal directions.

\begin{table}[t]
\caption{Percentage of errors on an MNIST rotated dataset \cite{mnist_rotated}.}
\label{MNIST_rotated}
\begin{center}
\begin{tabular}{|c c c c|}
\hline
PCA & PCA & PCA & Conv. \\
$m=1$ & $m=2$ & $m=3$ & Net. \\ 
\hline
$6.3$ & $3$ &  $\bf{2.8}$ & $8.8$ \\
\hline
\end{tabular}
\end{center}
\end{table}

The US-Postal Service dataset
is another handwritten digit 
dataset, with 7291 training samples
and 2007 test images $16 \times 16$ pixels.
The state of the art is obtained with 
tangent distance kernels \cite{tangent}.
Table \ref{usps} gives 
results obtained with the PCA classifier
and a polynomial kernel SVM classifier applied to 
scattering coefficients. The 
scattering scale was also set to $J=3$ by cross-validation. 

\begin{table}[t]
\caption{Percentage of errors for the whole USPS database.}
\label{usps}
\begin{center}
\begin{tabular}{|c c c c c|}
\hline
Tang. & SVM & PCA & PCA & PCA \\ 
Kern. & $m=2$ & $m=1$ & $m=2$ & $m=3$ \\ 
\hline
\textbf{2.4} & 2.64 & 3.24 & 2.74 & 2.74 \\
\hline
\end{tabular}
\end{center}
\end{table}

\subsection{Scattering Texture Classification}

\begin{figure}[ht]
\caption{The right plot gives scattering coefficients,
ordered according to their scattering order $q$. 
Blue coefficients correspond to $q=1$, green coefficients correspond to $q=2$ 
and red coefficients to $q=3$. Notice the exponential amplitude decay as
the order increases.}
\label{scattext}
\centering
\includegraphics[scale=0.26]{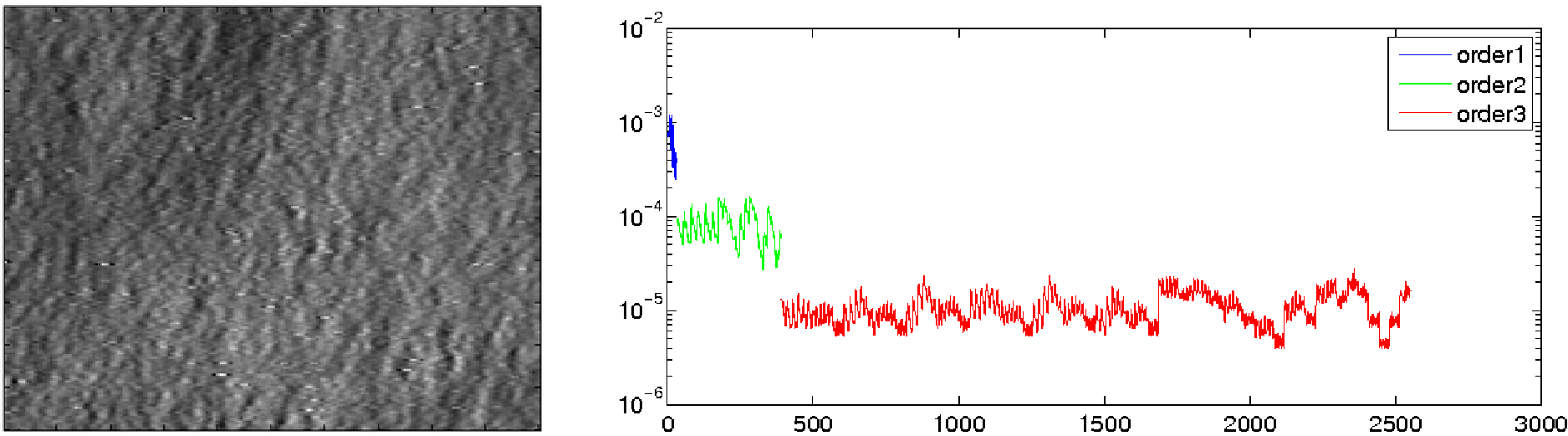}
%\subfigure[]{
%\includegraphics[scale=0.16]{cure1_im.eps}
%\label{fig:subfig1}
%}
%\subfigure[]{
%\includegraphics[scale=0.23]{scatt_cure1.eps}
%\label{fig:subfig2}
%}
%\subfigure[]{
%\includegraphics[scale=0.16]{cure4_im.eps}
%\label{fig:subfig3}
%}
%\subfigure[]{
%\includegraphics[scale=0.22]{scatt_cure4.eps}
%\label{fig:subfig4}
%}
\end{figure}

Scattering coefficients provide new texture descriptors, 
carrying co-occurrence information at different scales 
and orientations. A texture
can be modeled as a realization of a 
stationary process $F(x)$. Scattering coefficients $S_{J} F(x)$
are obtained with successive convolutions and modulus 
operators which preserve stationarity. Averaging by $\phi_J$ does not
modify expected values so 
$E \{S_{J} F(x)\}$ is a vector whose coefficients do 
not depend upon $x$ and $\phi_J$.
The convolution with $\phi_J$ reduces the coefficient
variability and for a large class of ergodic processes, 
the variance of $S_{J} F(x)$
decreases exponentially to zero as $J$ increases. As a result,
$S_{J} F(x)$ is a good estimator of 
$E \{S_{J} F (x)\}$ when $J$ is sufficiently large.
Figure \ref{scattext} shows an example of such vector for
a textured image with $m=3$.

\setcounter{subfigure}{0}

\begin{figure}[h!]
\centering
\caption{Left and right Brodatz textures $F_1$and $F_2$ 
have different power spectrum. The middle texture
$\widetilde F_1$ is obtained by filtering $F_1$
to equalize its power spectrum with $F_2$.}
\label{norm_textures}
\includegraphics[scale=0.3]{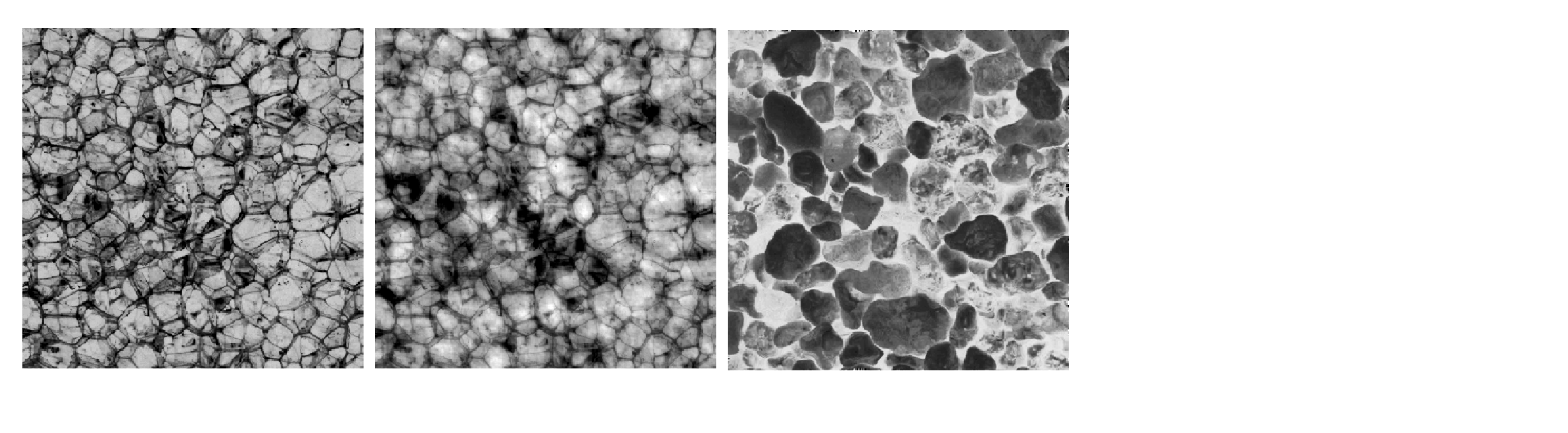}
%\subfigure[]{
%\includegraphics[scale=0.2]{brod_1n.eps}
%\label{fig:subfig1}
%}
%\subfigure[]{
%\includegraphics[scale=0.2]{brod_2n.eps}
%\label{fig:subfig2}
%}
%\subfigure[]{
%\includegraphics[scale=0.2]{brod_equalized.eps}
%\label{fig:subfig3}
%}
\end{figure}

Textures having same mean and same power spectrum have nearly the same 
scattering coefficients of order $q=0$ and $q = 1$. However, different textures
typically have co-occurence coefficients of order $q \geq 2$ which are
different. 
Let $S_{q,J} F_i$ be the vector of 
scattering coefficients of order $q$ for a texture
$F_i$. 
The distance of scattering vectors of order $q$ for
two textures $F_1$ and $F_2$ is normalized by
their variance $\sigma^2(S_{q,J} F_i)$:
\[
\rho_q(F_1,F_2) = \frac {|E \{S_{q,J} F_1\}-E \{S_{q,J} F_2\}|^2}
{\sigma^2(S_{q,J} F_1) + \sigma^2(S_{q,J} F_2)}~.
\]
Table \ref{results_brodatz} gives $\rho_q(F_1,F_2)$ for 
two Brodatz
textures in Figure \ref{norm_textures},
which have different power spectrum. 
Their expected scattering vectors $E \{S_J F_{q,i} \}$ have a relatively
large distance $\rho_q(F_1,F_2)$ at all orders $q \geq 1$. 
The texture $\widetilde F_1 $ in
Figure \ref{norm_textures} has same power spectrum as
$F_2 $. When $q = 1$, equalizing the power spectrum
reduces $\rho_q(\widetilde F_1,F_2)$ to
$0$ (up to estimation errors) but 
$\rho_q(\widetilde F_1,F_2)$ remains well above zero for $q > 1$.
Textures having same power
spectrum can thus be discriminated from scattering coefficients of
order $q > 1$.

\begin{table}[t]
\caption{Normalized distance $\rho_q$ of expected scattering vectors 
of order $q$, for textures in Figure \ref{norm_textures}.} 
\label{results_brodatz}
\begin{center}
\begin{tabular}{|c | c c|}
\hline
$q$ & $\rho_q(F_1,F_2)$ & $\rho_q(\widetilde F_1,F_2)$ \\ 
\hline
1 & 12 & 0 \\
2 & 12 & 1 \\
3 & 6 & 2 \\
4 & 3 & 2 \\
\hline
\end{tabular}
\end{center}
\end{table}

\begin{figure}[hb]
\centering
\includegraphics[scale=0.35]{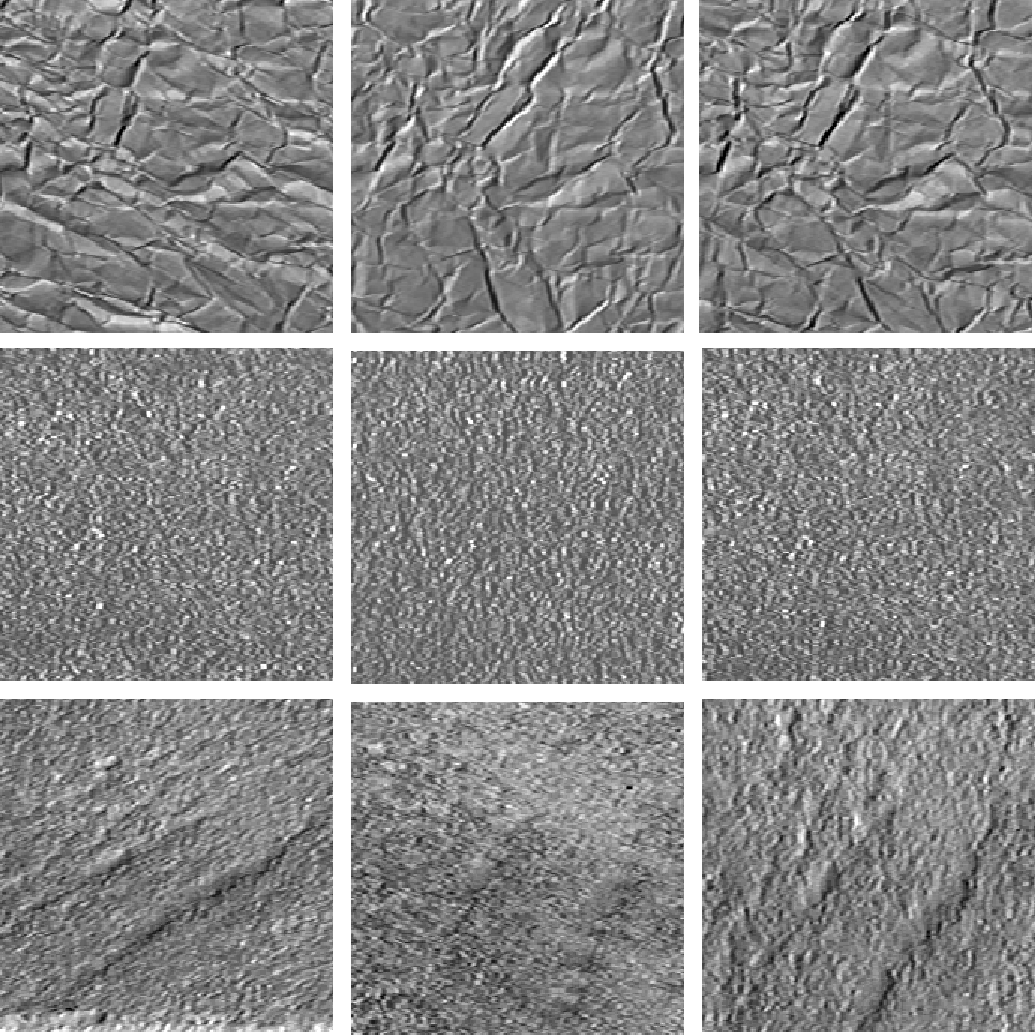}
%\caption{Top row: images of the same texture material with different poses and illuminations. Bottom row: examples of textures 
%that are in different classes despite their similarities.}
\caption{Examples of textures from the CUReT database. Each row corresponds to a different class, showing intra-class variability
in the form of stochastic variability and changes in pose and illumination.}
\label{curetfig}
\end{figure}

Texture classification is tested on 
the CUReT texture database \cite{Malik,Zisserman}, which includes
61 classes of image textures of $N = 200^2$ pixels.
%with 46 training samples and 46 testing samples in 
%each class. 
Each texture class gives images of
the same material with different
pose and illumination conditions. 
Specularities, shadowing and surface normal 
variations make it challenging for classification. 
Pose variations require global rotation invariance.
Figure \ref{curetfig} 
illustrates the large intra class variability, and also shows 
that the variability across classes is not always important.  

State of the art on this database achieves a
2.46\% error rate, obtained in \cite{Zisserman} with
an optimized Markov Random Field model.
The scattering PCA classifier has
a 0.09\% error rate, which is a factor 25 improvement,
as shown in Table \ref{curet}. 
The database is randomly split into a training
and a testing set, which either comprises 46 training images
each as in \cite{Zisserman}, or contains
23 training images as in \cite{curet_rotation}.
Results are averaged over 10 different splits.

\begin{table}[t]
\caption{Percentage of errors on CUReT for different training sizes.} 
\label{curet}
\begin{center}
\begin{tabular}{|c |  c c c c|}
\hline
Training  & PCA & SVM &LBP & MRFs  \\ 
size & $m=2$ &$m=2$& \cite{curet_rotation}  & \cite{Zisserman,curet_rotation} \\
\hline
23 &  $\bf{0.9 \pm 0.1}$ & 3.3 &18.23 & 22.43 \\
46 &  $\bf{0.09 \pm 0.05}$ & 1.1& 3.96 & 2.46 \\
\hline
\end{tabular}
\end{center}
\end{table}

The cross-validation adjusts the scattering scale $2^J = 2^7$
which is the maximum value. Indeed, these textures are
fully stationary and increasing the scale reduces the variance of 
the scattering coefficients variability across realizations.
Global invariance to rotation and illumination is provided by
the PCA affine space models. They include the main variation directions
of scattering vectors due to rotations or illumination variations.

The dimension of affine approximation space models is adjusted by
cross validation to $d=6$ and $d=22$ respectively for $23$ and $46$
training samples. The resulting error rates are respectively
$0.9\%$ and $0.09\%$. 
With an SVM using a polynomial kernel, the classification error 
for 46 training samples per class increases to
$1.1 \%$.
The intra class normalized approximation error 
$\sigma^2_d$ is only $2.5 \cdot 10^{-3}$ 
when using $46$ training samples,
about half of the error produced
 in the case of $23$ training samples, in 
 which $\sigma^2_d$ is $5.3 \cdot  10^{-3}$. 
 The estimated separation ratio is $\lambda_d = 8$ and
 $\lambda_d=5$ respectively. 
 Such low approximation errors are
 possible thanks to the fast variance decay
 of scattering coefficients as the scale increases
 and to the global invariance properties provided by 
 the affine spaces.

\section{Conclusion}

A scattering transform provides a locally translation invariant representation,
which linearizes small deformations, and provides co-occurrence coefficients
which characterize textures. 
For handwritten digit recognition and texture discrimination 
with small training size sequences, 
a PCA model selection classifier 
yields state of the art results.

Besides translations, invariance can be extended to any
compact Lie group $G$, by combining another scattering transform 
defined on $G$. The cascade of wavelet transforms in 
$\bf L^2( R^2)$ is then replaced by a cascade of wavelet transforms
in ${\bf L^2}(G)$ \cite{stephane1}.

\end{document}